\documentclass{article} 
\usepackage{iclr2020_conference,times}

\usepackage{amsmath,amsfonts,bm}

\def\eqref#1{equation~\ref{#1}}

\def\1{\bm{1}}

\DeclareMathAlphabet{\mathsfit}{\encodingdefault}{\sfdefault}{m}{sl}
\SetMathAlphabet{\mathsfit}{bold}{\encodingdefault}{\sfdefault}{bx}{n}

\usepackage{hyperref}
\usepackage{url}

\usepackage{epsfig}
\usepackage{cite}
\usepackage{transparent}
\usepackage{graphicx, subfigure}
\usepackage{float}
\usepackage{amsmath}
\usepackage{amssymb}
\usepackage{booktabs, multirow}
\usepackage{bm}
\usepackage{enumerate,enumitem}
\usepackage[ruled,linesnumbered]{algorithm2e}

\title{A New Framework for Query Efficient Active Imitation Learning}

\author{Daniel Hsu \\
Department of Electrical and Computer Engineering\\
Georgia Institute of Technology\\
Atlanta, GA 30318, USA
}

\begin{document}

\maketitle

\begin{abstract}
We seek to align agent policy with human expert behavior in a reinforcement learning (RL) setting, without any prior knowledge about dynamics, reward function, and unsafe states. There is a human expert knowing the rewards and unsafe states based on his preference and objective, but querying that human expert is expensive. To address this challenge, we propose a new framework for imitation learning (IL) algorithm that actively and interactively learns a model of the user’s reward function with efficient queries. We build an adversarial generative model of states and a successor feature (SR) model trained over transition experience collected by learning policy. Our method uses these models to select state-action pairs, asking the user to comment on the optimality or safety, and trains a adversarial neural network to predict the rewards. Different from previous papers, which are almost all based on uncertainty sampling, the key idea is to actively and efficiently select state-action pairs from both on-policy and off-policy experience, by discriminating the queried (expert) and unqueried (generated) data and maximizing the efficiency of value function learning. We call this method adversarial reward query with successor representation. We evaluate the proposed method with simulated human on a state-based 2D navigation task, robotic control tasks and the image-based video games, which have high-dimensional observation and complex state dynamics. The results show that the proposed method significantly outperforms uncertainty-based methods on learning reward models, achieving better query efficiency, where the adversarial discriminator can make the agent learn human behavior more efficiently and the SR can select states which have stronger impact on value function. Moreover, the proposed method can also learn to avoid unsafe states when training the reward model.    
\end{abstract}

\section{Introduction}
In reinforcement learning problem, the objective of RL agent is always set by reward functions \citep{sutton1998introduction}. Such reward functions are usually hand coded and, unfortunately, defining rewards for real robot tasks manually is challenging even for relatively well-understood problems such as robotic grasping. Recent learning from demonstrations (LfD) \citep{argall2009survey} algorithms aim to provide an intuitive interface that allows end-users to program robots without the use of code or expert knowledge. Generative Adversarial Imitation Learning (GAIL) \citep{ho2016generative} is a form of LfD that focuses on recovering the underlying reward function that generates an expert’s demonstrations \citep{abbeel2004apprenticeship}. Reinforcement learning in conjunction with this optimized reward function always shows better generalization capability, compared with supervised policy learning, such as Behavior Cloning \citep{bain1999framework}, because GAIL captures the underlying intention of the demonstrator and attenuates the compounding error caused by covariate shift \citep{ross2010efficient}.

A significant problem for existing IL algorithms is that they often require a large number of demonstrations from human expert, while it is also difficult for human expert to determine what kinds of demonstrations are most informative to train the RL agent. To deal with this problem, active IL \citep{lopes2009active, cohn2011comparing, cui2018active} have been proposed that estimate the uncertainty and information gain to select queries that are expected to be the most informative under certain criteria. However, previous papers on active IL are all based on uncertainty sampling. Uncertainty estimation can be difficult in problems with high-dimensional observations and complex state dynamics. Besides, diversity is always neglected in query selection by these papers, which may lead to select similar query states and reduce efficiency. Moreover, uncertainty-based methods are vulnerable to outliers. 

In this paper, we propose an active IL method which learns a low dimensional latent space from queried and unqueried data using Wasserstein Autoencoders (WAE) \citep{tolstikhin2017wasserstein}. Based on this learned latent space, we propose two methods to select data from the unqueried pool for human labeling. The first method selects data that are sufficiently different in the latent space from queried data. It is performed by an adversarial network which classifies which pool the data instances belong to (queried or unqueried) and does not depend on the task or policies for which are trying to collect human labels. The second method selects {\it bottleneck} states for human labeling, which are prototypical representatives of well-connected regions in state space and can hence have stronger impact on learning value function than other states. Instead of constructing a graph of the state space, it leverages Successor Representations (SR) \citep{dayan1993improving} to learn the bottleneck states. Since the temporal structure among states can be captured by the SR inherently, a reasonable proxy for the actual graph can be formed implicitly. By using the latent space learned by WAE as state features, the SR can be learned with temporal-difference (TD) learning \citep{sutton1998introduction} and, as we discuss below, it can be seen as implicitly estimating the transition dynamics of the environment. Moreover, some recent papers have integrated SR with neural network function approximators \citep{kulkarni2016deep,barreto2017successor}, allowing us to implicitly form the graphical structure of high-dimensional state space in an efficient way. 

\section{Preliminaries}
In this section, we provide a review of some established models adopted in this paper. 
\subsection{Generative Adversarial Imitation Learning}
The base model of imitation learning (IL) in this paper is Generative Adversarial Imitation Learning (GAIL) \citep{ho2016generative}. GAIL is the-state-of-art IL model which has shown success in many difficult problems. It builds on GAN \citep{goodfellow2014generative}, which is a popluar generative model and shows success in many real-world tasks. GAN uses two networks, a generator $G$ and a discriminator $D$. The generator tries to cheat the discriminator and generate samples that are indistinguishable from real data. The job of the discriminator is then to tell apart the data and the samples, predicting 1 with high probability if the sample is real and 0 otherwise. More specifically, GANs optimize the following objective function
\begin{equation}
    \min_G\max_D\mathbb{E}_{x\sim\mathcal{D}_{\text{data}}}[\log D(x)] + \mathbb{E}_{p(z)}[\log(1-D(G(x)))] \nonumber
\end{equation}

Inspired from GAN \citep{goodfellow2014generative}, GAIL is a model-free adversarial imitation learning framework. The reward given by the discriminator $D$ drives the policy $\pi$ to imitate the expert (human) policy behavior $\pi_{\text{E}}$ by minimizing the Jensen-Shannon divergence between the state-action distributions generated by $\pi$ and the expert state-action distribution by $\pi_{\text{E}}$ via the learning objective
\begin{equation}
\min_{\pi}\max_{D\in(0,1)^{\mathcal{S}\times\mathcal{A}}}\mathbb{E}_{\pi}\big[\log D(s,a)\big]+\mathbb{E}_{\pi_{\text{E}}}\big[\log(1-D(s,a))\big]-\lambda H(\pi) \label{gail} 
\end{equation}
where $\lambda$ is a hyper-parameter and $H(\pi)$ is the entropy regularization term for policy $\pi$. Here the discriminator $D$ performs the binary classification to distinguish between samples generated by learning policy $\pi$ and target policy $\pi_{\text{E}}$. 

Behavior Cloning (BC) \citep{pomerleau1991efficient,rusu2015policy,duan2017one} is another popular imitation learning method. It can perform well when expert demonstrations are plentiful. However, the performance of BC can degenerate without abundant data \citep{ross2010efficient, ross2011reduction}. When using BC, even small errors in mimicking the expert demonstration behavior can be quickly accumulated as the policy is unrolled. A good policy should correct for mistakes made previously, but for BC to achieve this, the corrective behaviors have to appear frequently in the training data.

\subsection{Wasserstein Autoencoder}
In order to learn generative models and latent representation for high-dimensional observations in IRL problems, we adopted the recently proposed Wasserstein Autoencoder (WAE) \citep{tolstikhin2017wasserstein,rubenstein2018latent}. VAE \citep{kingma2013auto} is a dominant model in learning latent representation learning and generative modeling, which focuses on minimizing KL divergence. But it can produce samples which are far from the true data manifold, especially true for structured  high-dimensional  datasets  such  as  natural images. Different from popular models such as VAE \citep{kingma2013auto}, WAEs switche to minimize the optimal transport distance between the input data distribution $P_X$ and output generation distribution $P_G$. Given any non-negative cost function $c(x,x')$ between two input instances, denoting latent code as $Z$ in latent space $\mathcal{Z}$, WAEs minimize the following objective in terms of parameters of the encoder $Q(Z|X)$ and decoder $G(X|Z)$,
\begin{equation}
    \min_{Q, G}\mathbb{E}_{P_X}\mathbb{E}_{Q(Z|X)}[c(X, G(Z))] + \beta D_Z(Q_Z, P_Z) \label{wae}
\end{equation}
where $P_Z$ is the prior distribution of latent code $Z$, and $Q_Z:=\int Q(Z|X)P(X)dX$ is the aggregated posterior distribution, $D_Z$ is any divergence in latent space $\mathcal{Z}$ and $\beta > 0$ is the regularization hyper-parameter. In this paper, we find an adversarial divergence works well in image-based imitation learning problems.

\subsection{Successor Representation}
The Successor Representation (SR) represents a state in terms of its successors \citep{dayan1993improving}. Originally, the SR for $s$ is defined as a vector of $|\mathcal{S}|$ with $i$-th index equal to the discounted future occupancy for state $s_i$ given the agent starting at $s$. Since the SR counts the visitation of successor states, it is directly dependent on the policy $\pi$ and the transition dynamics $p(s_{t+1}|s_t, a_t)$. Specifically, the original definition of SR can be written as 
\begin{equation}
    \varphi_{\pi}(s,s')=\mathbb{E}_{s'\sim P, a\sim\pi}\bigg[\sum_{t=0}^{\infty}\gamma^t\mathbb{I}(s_t=s')\bigg|s_0=s\bigg] \label{sr0}
\end{equation}
where $\mathbb{I}$ denotes the indicator function. The SR can also be updated by a temporal difference (TD) \citep{tesauro1995temporal} approach by writing it in terms of the SR of the next state:
\begin{equation}
    \hat{\varphi}(s_t, \cdot):=\hat{\varphi}(s_t,\cdot) + \alpha\bigg[\mathbf{1}_{s_t}+\gamma[\hat{\varphi}(s_{t+1},\cdot)]-\hat{\varphi}(s_t,\cdot)\bigg] \label{sr1}
\end{equation}
where $\hat{\varphi}$ is the estimated SR being updated, and $\mathbf{1}_{s_t}$ is the one-hot vector of state $s_t$ with respect to all other states. However, the original definition of SR cannot be applied to MDP with high-dimensional states, where the number of all states is overwhelming. Assuming state is represented by $K$-dimensional features $\phi(s)$, some recent papers \citep{kulkarni2016deep, barreto2017successor} extent SR to high-dimensional problems by using deep neural network as approximators
\begin{equation}
    \varphi_{\pi}(s_t;\theta)=\mathbb{E}[\phi(s_t)+\gamma\varphi_{\pi}(s_{t+1};\theta)] \label{sr2} 
\end{equation}

\section{Methodology}
In this section, we are going to introduce the proposed model and query selection strategies. Here we propose both on-policy and off-policy query strategies. The on-policy strategy is to select state-action pairs from the learning policy which are different enough from the expert policy, so as to avoid unsafe actions during exploration. The off-policy strategy is to pick up states that are bottleneck states or form the core-set from the large amount state-action pairs in the replay buffer, so that the value function can be updated more efficiently. Here we denote $\phi(s_t)$ as the latent of input state $s_t$, and $\varphi(s_t)$ as the successor representation of state $s_t$.

\subsection{Model}
The base model is AIRL \citep{fu2017learning}, a state-of-art model for IRL achieving many success in large-scale problems. Different from previous work, here we consider the active IRL in large-scale problems with high-dimensional observations. So, we introduce a Wasserstein Autoencoder (WAE) to learn a low-dimensional latent space of the input states, and train the discriminator and successor representation on that latent space. Besides, for large-scale image-based IRL problems, in addition to the divergence regularizing term in regular WAE objective \eqref{wae}, we introduce another adversarial regularization term, measuring the distance in the learned latent space between the state-action pairs collected by the learning policy and expert policy.

 \begin{figure}[H]
    \def\svgwidth{\columnwidth}
    \centering    
    \scalebox{0.5}{\large{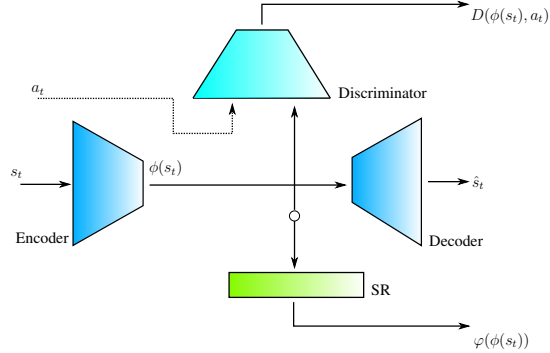}}
    \caption{The proposed model. The empty circle here denotes the stop gradient during training. SR is short for successor representation.}
    \label{fig:model}
\end{figure} 

We use the same definition of discriminator $D$ as \eqref{airl}. Regarding WAE \citep{tolstikhin2017wasserstein}, based on empirical study, we choose to use deterministic encoder. Specifically, in \eqref{wae}, the prior of latent codes $P_Z$ is unit multivariate Gaussian $\mathcal{N}(0, \mathbb{I})$. The cost function $c(S, S')$ is chosen to be l-2 norm loss. We choose the divergence distance $D_Z$ to be maximum mean discrepancy (MMD) \citep{gretton2012kernel}, where the Radial Basis Function (RBF) kernel and Rational Quadratic (RQ) Kernel are both evaluated. Denote the encoder output of WAE as $\phi(s)$ and decoder output as $G(z)$, the training objective of WAE is as below
\begin{equation}
    \mathcal{L}_{\text{WAE}}(\mathcal{D}_{S}, \mathcal{D}_{Z})=\sum_{(s, z)\in\mathcal{D}_S\cup\mathcal{D}_Z}\|s, G(\phi(s))\|_2+\beta_1 \widetilde{\text{MMD}}(\phi(\mathcal{D}_S), \mathcal{D}_Z) \label{wae_o}
\end{equation}
where the sets $\mathcal{D}_S$ and $\mathcal{D}_Z$ must have the same size, and $\widetilde{\text{MMD}}(\cdot, \cdot)$ is the empirical MMD where integral is approximated by discrete summation \citep{tolstikhin2017wasserstein}. In addition, we find an adversarial regularizer can improve the practical performance a lot, which is the empirical MMD between state latent from learning policy and expert policy, i.e., $\widetilde{\text{MMD}}(\phi(\mathcal{D}_S^P), \phi(\mathcal{D}_S^E))$, where $\mathcal{D}_S^P$ and $\mathcal{D}_S^E$ are states sampled from replay buffer and expert demonstrations, and they have the same size. Therefore, assuming the sets $\mathcal{D}_S, \mathcal{D}_A$ and $\mathcal{D}_{S'}$ denote current states, actions, and next states given the transitions sampled from learning policy and expert demonstrations, fixing learning policy $\pi$, the overall training objective of the proposed model is to minimize the loss as below:
\begin{eqnarray}
\lefteqn{\mathcal{L}((\mathcal{D}_S^P, \mathcal{D}_A^P), (\mathcal{D}_S^E, \mathcal{D}_A^E))} \nonumber \\
&=&{\sum_{(s,a)\in\mathcal{D}^P_S\cup\mathcal{D}^P_A}\log D(\phi(s),a) + \sum_{(s,a)\in\mathcal{D}^E_S\cup\mathcal{D}^E_A} \log(1-D(\phi(s),a)) } \nonumber \\
&&+ \alpha_1\mathcal{L}_{\text{WAE}}(\mathcal{D}_S^P, \mathcal{D}_A^P) +  \alpha_2\mathcal{L}_{\text{WAE}}(\mathcal{D}_S^E, \mathcal{D}_A^E) +  \beta\widetilde{\text{MMD}}(\phi(\mathcal{D}_S^P), \phi(\mathcal{D}_S^E)) \label{obj}
\end{eqnarray}
where the parameters of encoder, decoder and discriminator are all omitted. Moreover, the SR is learned by TD updates, whose training loss is defined as,
\begin{equation}
    \mathcal{L}_{\text{SR}}(\mathcal{D}_S^P, \mathcal{D}_{S'}^P)=\sum_{(s,s')\in\mathcal{D}_S^P\cup\mathcal{D}_{S'}^P}\|\varphi(s) - (\phi(s) + \gamma\varphi'(s'))\|_2 \label{sr_o}
\end{equation}
where $\varphi'$ is the target network of SR, and it is updated with $\varphi$ periodically. Note that the encoder, or state feature network, $\phi$, is not updated during the training of SR.

\subsection{On-policy Query: Safe Exploration}
Our first query strategy is to select states which are sufficiently different in latent space compared with those in expert dataset, to maximize the performance of the latent representation learned on the newly queried data. We aim to choose states whose optimality are uncertain enough and are not well represented in the queried (expert) dataset. It is agnostic to the imitation learning task. This selection for query has to be performed by an adversarial network. So we can directly utilize the discriminator in GAIL, trained by same number of transitions from learning policy $\pi$ and expert policy $\pi^*$. 

Fortunately, it can be also regarded as a safety classifier \citep{zhang2017query}, returning label indicating whether the action given by the learning policy $\pi$ is likely to deviate from that by the expert policy $\pi^*$ without querying it. So we can also have safe exploration while conducting this query selection in an on-policy way. However, the safety classifier in \citep{zhang2017query} is trained by binary logistic regression. Here the discriminator $D$ is trained in an adversarial way. We are going to show that the adversarial learning approach is more query-efficient to learn safety strategy than that in \citep{zhang2017query}. 

The safe strategy adopted here is that at every time point, the discriminator $D$ determines if it is safe to take the action from the leaning policy $\pi$ given current state. In training iterations, we can maintain a threshold $\tau$, measuring the safety of the state-action pair $(s,a)$ suggested by the learning policy $\pi$. If $D(\phi(s),a)<\tau$, it shows that $(s,a)$ is too different from that the expert behavior and it is not safe to take action suggested by $\pi$. In this case, the agent will query the expert and take the action from $\pi^*(a'|s)$. If $D(\phi(s),a)>\tau$, it is safe to follow $\pi$ and take the action $a$. 

The threshold $\tau$ can be fixed initially, tuned empirically in specific environments. It can also be updated in every iteration. For example, by the end of every iteration, $\tau$ can be set to the $\alpha$-quantile of the discriminator outputs of all state-action pairs taken by $\pi$ in this iteration. This $\alpha$ can be very small, such as $0.02$ or $0.05$, so that the selected states for query are sufficiently different from those in expert dataset.
    
\subsection{Off-policy Query: A Core-set on Successor Representations}
Our second query strategy is to select {\it landmark} states, which are that are representatives of well-connected regions in the state space. These states always have stronger impact on value function learning than others. Inspired by the batch acquisition for active learning \citep{sener2017active,pinsler2019bayesian,kirsch2019batchbald}, we propose a core-set approach for query selection, which aims to find a small subset of the whole unqueried states in the replay buffer, so that the value function learned over the subset of states is competitive over the whole set of unqueried states. 

Here we formulate the query selection as a K-center problem \citep{lim2005k, sener2017active}, which is to cluster states and find the core-set states based on the learnt successor representations (SR) \citep{kulkarni2016deep, barreto2017successor} of the unqueried states in replay buffer. This process is conducted in an off-policy way. Here the core-set is found by k-medians algorithm for computational simplicity. Since the SR captures temporally close-by states efficiently, the generated clusters are spread across the state space, with each cluster assigned to a set of states that are densely connected.

\subsection{Overall Algorithm}
The whole algorithm is shown as below. In the beginning, the expert dataset is initialized with state-action pairs from human demonstration behavior. In order to show the effect of active queries, the expert dataset can only have small number of data instances, or even be empty. The update of threshold $\tau$ in on-policy query can be periodical and specific to the exact environment. For example, we can choose $\tau$ to cover the lowest 5\% discriminator scores of state-action pairs in the replay buffer. The on-policy query is conducted at Line 6, and the off-policy query is from Line 15 to 16. Empirically, we find both on-policy and off-policy queries cannot be conducted too frequently. Because queries can change the expert dataset, which may negatively influence the training stability of GAIL.
\begin{algorithm}
\caption{Query-efficient Active Imitation Learning}
\label{alg1}
Initialize expert dataset $\mathcal{E}$, and set replay buffer $\mathcal{B}$ to be empty. Initialize the parameters of all models. Initialize state $s$. Initialize safety threshold $\tau\leftarrow0$. \;
Input the off-policy query interval $T_{\text{Off}}$ and number of centers in clustering $N_K$. \;
\For{$t=1,2,\ldots$}
{
Get action $a$ from learning policy $\pi(\cdot|s)$. \; 
\If{$D(\phi(s),a)<\tau$}{Ask human expert for the optimal action $a^*$ in state $s$. \; 
Set $a\leftarrow a^*$}
Apply the action $a$, and get the next state $s'$. \;
$\mathcal{B}\leftarrow\mathcal{B}\cup(s,a,s')$. \;
Sample minibatch $(\mathcal{D}^E_S, \mathcal{D}^E_A)$ and $(\mathcal{D}^P_S, \mathcal{D}^P_A)$ from $\mathcal{E}$ and $\mathcal{B}$ respectively. \;
Update policy $\pi$ with rewards in terms of discriminator output $D$, i.e., $\hat{r}=\log D-\log(1-D)$. \;
Update the WAE and discriminator by minimizing the total loss $\mathcal{L}$ in \eqref{obj} with $(\mathcal{D}^E_S, \mathcal{D}^E_A)$ and $(\mathcal{D}^P_S, \mathcal{D}^P_A)$\;
Update the SR by minimizing \eqref{sr_o} with  $(\mathcal{D}^P_S, \mathcal{D}^P_A)$. \;
\If{$t \mod T_{\text{Off}} == 0$}{Conduct $N_K$-median clustering over $\{\varphi(s)|\forall s\in\mathcal{B}\}$, and yield centers $\{s^k\}_{k=1}^{N_K}$ \;
Ask human expert for the optimal actions $a^*$ at $\{s^k\}_{k=1}^{N_K}$, and update expert dataset $\mathcal{E}\leftarrow\mathcal{E}\cup\{(s^k, a^{*k})\}_{k=1}^{N_K}$. }
}
\end{algorithm}

\section{Experiment}
\subsection{Maze Game}
We first evaluate the proposed algorithm on maze, which is a 2D navigation game. The map divided into 10x10 equal-sized blocks. The map and policy loss learning curve are shown in Figure \ref{fig:maze}. The agent and target occupy one block. The agent can start in any empty grid and the target is at the lower-right corner of the map. The action space consists of four directional movements. The maze architecture is kept fixed for the purpose of this work. If an agent moves, but hits a wall, it will stay in the original grid. The rewards in this game is 10 for reaching the target and -1 for every movement. The benchmarks are random query and uncertainty-based query. The former is to randomly select state-action pairs for query from replay buffer. The latter is based on uncertainty estimation of the Q-value of state action pairs. The model builds on bootstrapped DQN \citep{osband2016deep}, where the number of heads $K$ is set to 10. The expert dataset only has one expert trajectory initially.

\begin{figure}[H]
\subfigure[Map]{
\begin{minipage}[t]{0.45\linewidth}
\centering
\includegraphics[width=0.54\columnwidth]{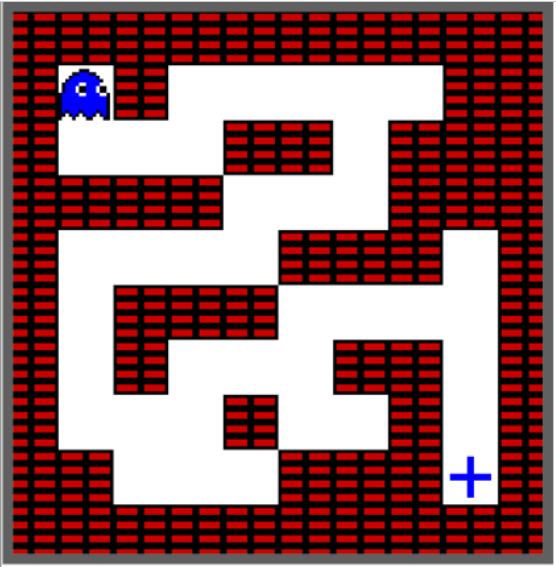}
\end{minipage}
}
\subfigure[Learning Curve]{
\begin{minipage}[t]{0.45\linewidth}
\centering
\includegraphics[width=0.82\columnwidth]{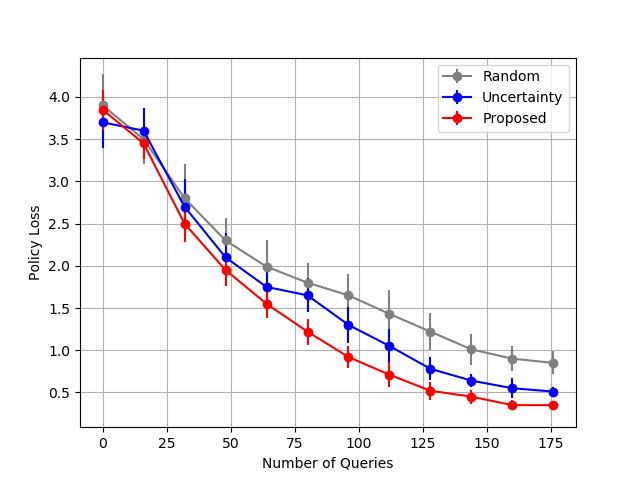}
\end{minipage}
}
\caption{The proposed method evaluated on 2D Navigation Maze Game.}
\label{fig:maze}
\end{figure}

\subsection{Car Racing Game}
We also evaluated the proposed algorithm on a 2D driving game, to test its performance on tasks with high-dimensional states. Here the state $s\in\mathbb{R}^{96\times96\times3}$ is a top-down view of the car. The action $a\in\mathbb{R}^3$ controls gas, steering and break. In each query, the expert returns to the RL agent the optimal action according to human behavior under current state. The benchmarks are same as the section above.

Regarding the architecture, since this game has high-dimensional states, we use WAE to learn the latent representation of the state. The architecture of encoder and decoder of WAE are chosen to be same as that in \citep{barreto2017successor}. The SR is learned by using the encoder output of WAE as state features.

\begin{figure}[H]
\subfigure[Screen]{
\begin{minipage}[t]{0.45\linewidth}
\centering
\includegraphics[width=0.72\columnwidth]{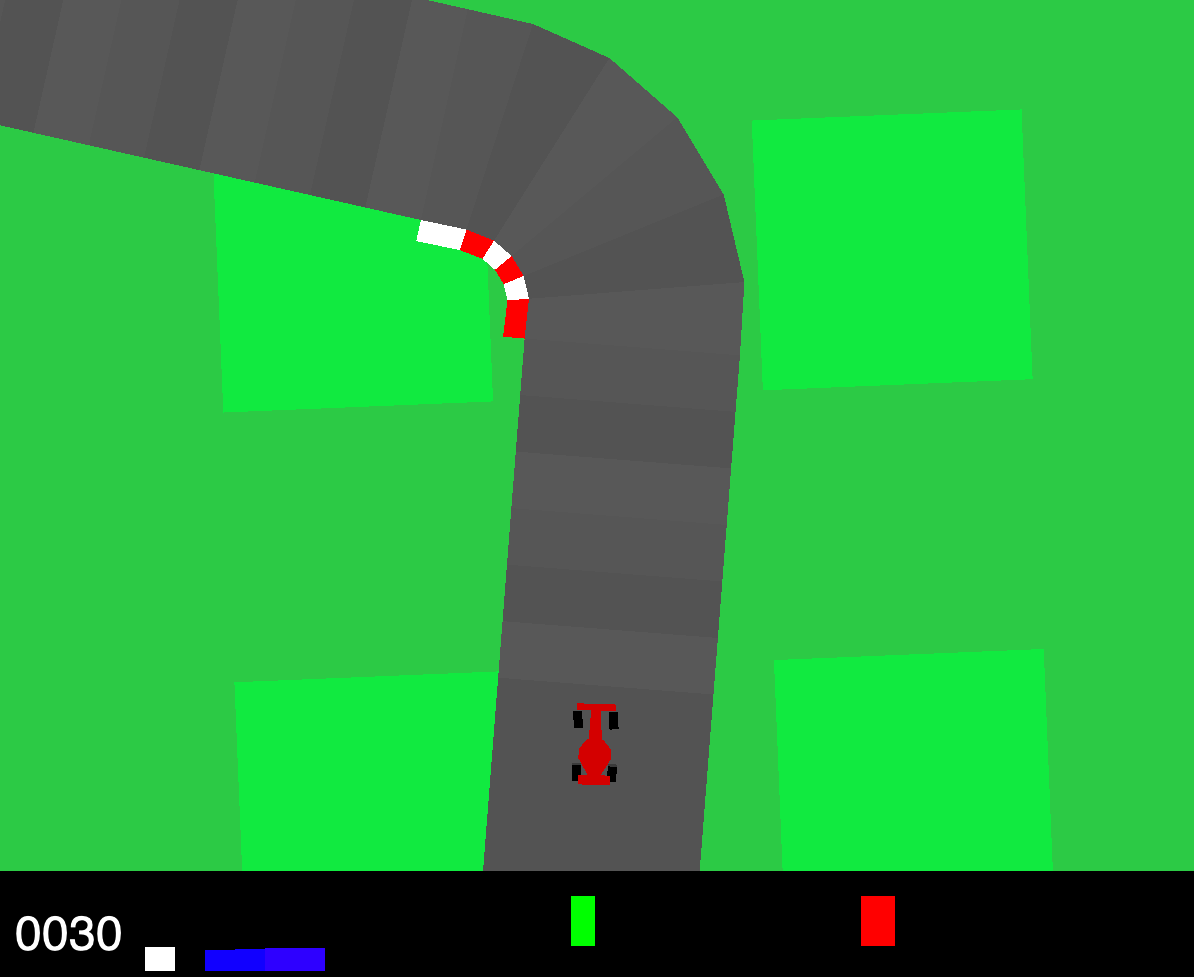}
\end{minipage}
}
\subfigure[Learning Curve]{
\begin{minipage}[t]{0.45\linewidth}
\centering
\includegraphics[width=0.89\columnwidth]{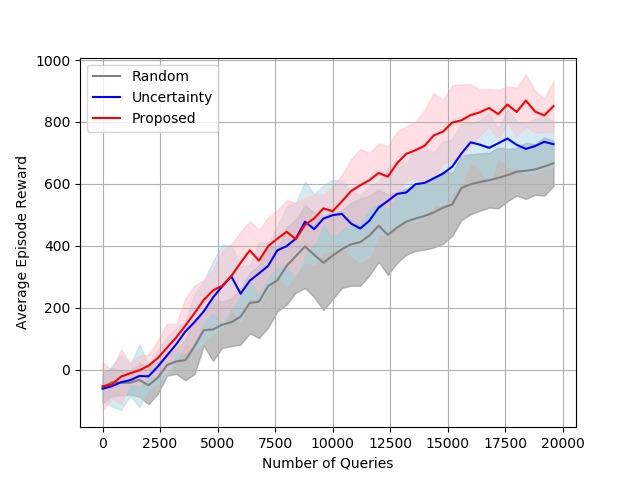}
\end{minipage}
}
\caption{The proposed method evaluated on Car Racing Game.}
\label{fig:maze}
\end{figure}

\section{Conclusion}
In this paper, we propose a new framework for imitation learning. It is based on two novel query selection strategies, on-policy and off-policy query. The first one builds on the adversarial network, which selects states that are sufficiently different from the queried states stored in expert dataset. The other utilizes the representation power of successor representation (SR), which can pick most representative states on policy's occupancy measure and improve the efficiency of learning of value function. The evaluation on practical games validates the advantage of the proposed algorithm.

\bibliography{rl}
\bibliographystyle{iclr2020_conference}

\end{document}